\documentclass[journal,article,pdftex,moreauthors,preprints,accept]{Definitions/mdpi} 

\firstpage{1} 
\pubvolume{1}
\issuenum{1}
\articlenumber{0}
\pubyear{2023}
\copyrightyear{2023}
\datereceived{ } 
\daterevised{ } 
\dateaccepted{ } 
\datepublished{ } 
\hreflink{https://doi.org/} 

\usepackage{multirow}
\Title{Proximity-aware Self-attention-based Sequential Location
Recommendation}

\TitleCitation{Title}

\Author{Xuan Luo $^{1, *}$\orcidA{}, Mingqing Huang, Rui Lv $^{1}$ and Hui Zhao $^{1}$ }

\AuthorNames{Firstname Lastname, Firstname Lastname and Firstname Lastname}

\AuthorCitation{Lastname, F.; Lastname, F.; Lastname, F.}

\address{%
$^{1}$ \quad Shenzhen Institute of Information Technology, Shenzhen 518172, China;  \\
}

\corres{Correspondence: luoxuan@sziit.edu.cn; 
}

\abstract{Sequential location recommendation plays a huge role in modern life, which can enhance user experience, bring more profit to businesses and assist in government administration. Although methods for location recommendation have evolved significantly thanks to the development of recommendation systems,  there is still limited utilization of geographic information, along with the ongoing challenge of addressing data sparsity. In response, we introduce a Proximity-aware based region representation for Sequential Recommendation (PASR for short), built upon the Self-Attention Network architecture. 
We tackle the sparsity issue through a novel loss function employing importance sampling, which emphasizes informative negative samples during optimization. Moreover, PASR enhances the integration of geographic information by employing a self-attention-based geography encoder to the hierarchical grid and proximity grid at each GPS point. To further leverage geographic information, we utilize the proximity-aware negative samplers to enhance the quality of negative samples.
We conducted evaluations using three real-world Location-Based Social Networking (LBSN) datasets, demonstrating that PASR surpasses state-of-the-art sequential location recommendation methods.
}

\keyword{Sequential Recommendation; Geo-Hash; Proximity Region; Self-Attention; Encoder.} 

\begin{document}




\section{Introduction}

In the era of rapid information technology advancement, the process of digitizing and sharing human mobility behaviors with friends has become significantly streamlined. These mobility patterns offer valuable insights into understanding and forecasting human movements \cite{gonzalez2008understanding, wu2019graph}, thereby enhancing various aspects of daily life such as dining, transportation, and entertainment. However, the predictability of individual mobility remains a challenge \cite{lian2014analyzing, song2010limits} due to data gaps and sparsity. To predict a personalized ranking of locations based on an individual's mobility history, sequential location recommendation assumes a pivotal role in enhancing the predictability of human movement across unfamiliar places. This is achieved by harnessing collective insights. Beyond its impact on mobility prediction, sequential location recommendation finds utility across a spectrum of applications, including route planning and location-targeted advertising.

Recently, the techniques employed for sequential location recommendation have witnessed a notable evolution, progressing from matrix factorization and metric learning to the utilization of RNN/CNN-based neural networks. To illustrate, the expansion of Factorizing Personalized Markov Chains (FPMC) \cite{rendle2010factorizing} was undertaken to address the challenge of sparse representation in modeling personalized location transitions \cite{cheng2013you,lian2013collaborative}. In the domain of metric learning, the introduction of Personalized Ranking Metric Embedding (PRME) was intended to capture individualized patterns of location transitions \cite{feng2015personalized}. This concept was subsequently extended to encompass geographic influence by incorporating the product of travel distance and estimated transition probability. For capturing long-term dependencies, the incorporation of Recurrent Neural Networks (RNNs) like GRU and LSTM has been extended to integrate spatial-temporal information \cite{cui2019distance2pre,li2018next,liu2016predicting,wu2019graph}. This was achieved by embedding parameters such as travel distance, travel time, and so on. Moreover, the design of spatial-temporal gates to better utilize the spatial-temporal information has also been explored.

Among the prevailing methods, two noteworthy challenges remain inadequately addressed. Firstly, the effective utilization of geographic information continues to be a gap. It is widely acknowledged that the GPS coordinates of a location play a vital role in illustrating the distance between locations. Additionally, a user's historical mobility data often demonstrates a propensity for spatial clustering \cite{lian2014geomf,ye2011exploiting}. Thus, the accurate encoding of GPS positions becomes indispensable. Secondly, these methods might grapple with the issue of sparsity. It's important to highlight that a user typically frequents a finite set of locations \cite{song2010modelling}, leading to a mixture of negatively-preferred and potentially positive locations within individual unvisited places \cite{HeLHT23}. Current methods utilize either the BPR loss \cite{rendle2012bpr} or the binary cross-entropy (BCE) loss for optimization, by contrasting visited locations with randomly selected samples from unvisited ones. However, each sample's level of informativeness varies, making it clear that treating all negative samples equally in these functions falls short of the optimal approach.
\begin{figure}[t]
\centering
\includegraphics[width=14cm]{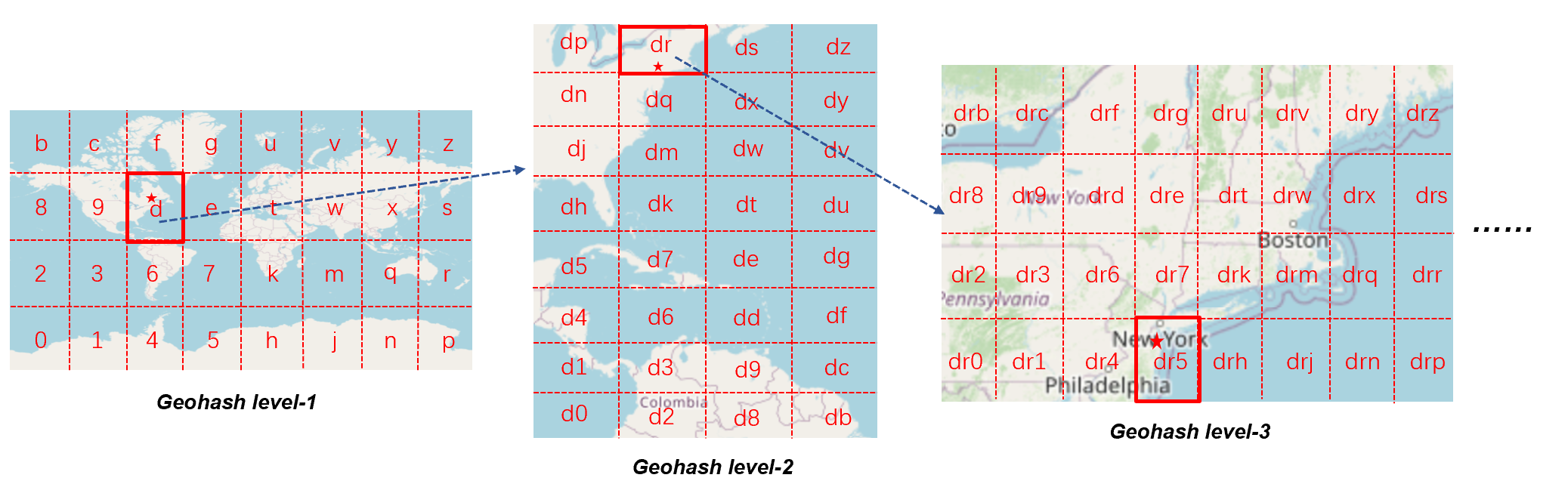}
\caption{Geohash example}
\label{fig_geohash}
\end{figure}  

\begin{figure}[t]
\centering
\includegraphics[width=12.5cm]{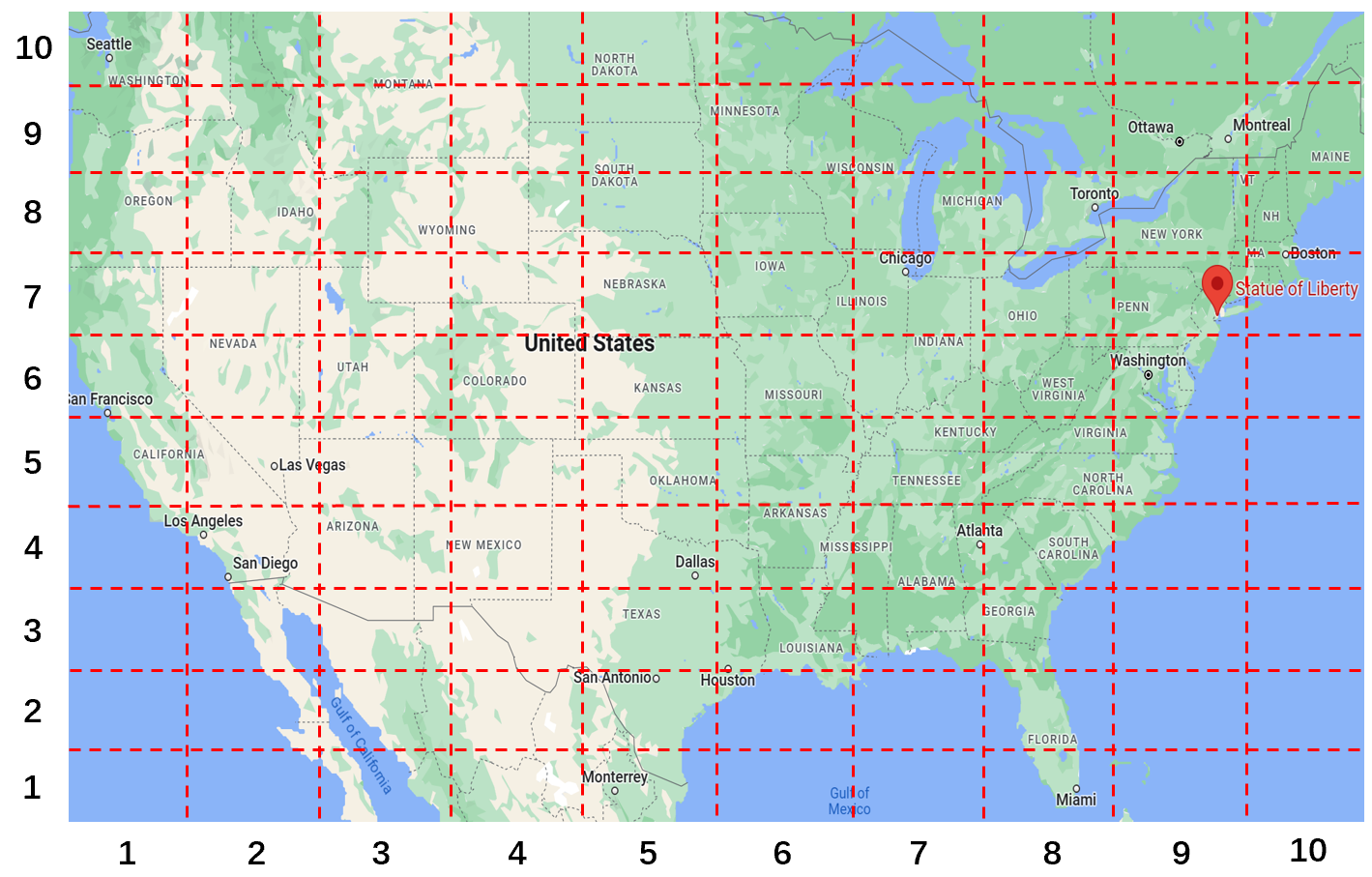}
\caption{Grid partition}
\label{fig_grid_sketch}
\end{figure}

To address these challenges, we introduce a Proximity-aware based region representation for sequential recommendation, referred to as PASR, built upon the foundation of the Self-Attention Network. PASR aims to enhance sequential location recommendation. In addition to incorporating location embeddings, PASR introduces a novel geography encoder and  a grid mapper to make better use of geography information. The geography encoder first utilizes the Geohash to encode a  geographic coordinate to a string of letters and digits. For instance, the GPS coordinates (40.68925, -74.0445) of the Statue of Liberty are converted into a geohash string 'dr5r7p62n1' using the Geohash algorithm, with a string length specified as 12. In \textbf{Figure} \ref{fig_geohash}, we give the first three layers of Geohash string w.r.t. the Statue of Liberty. Embedding the geohash string directly might seem straightforward, but it fails to capture complex spatial relationships. Note that the geohash strings of adjacent coordinates are similar and share some substrings. Therefore, we leverage the self-attention based encoder to encode the n-gram sequences of geohash strings. To enhance the proximity among positions, we propose to partition the region into grids. Concretely, The grid mapper divides the map into grids along latitude and longitude. Each grid is indexed by the row ID and column ID. In \textbf{Figure} \ref{fig_grid_sketch}, we give the grid partition as example where the ID pair of the Statue of Liberty is $(7,9)$. The concatenation of corresponding row embedding and column embedding make up the representation of the grid.

In response to the issue of sparsity, we introduce a novel weighted BCE loss employing importance sampling. This strategy assigns greater weight to informative negative samples. They play a more significant role in influencing the gradient by assigning more weight to informative samples. As a result, the gradient's magnitude increases, thereby expediting the training process. Notably, this adaptive loss function can be seamlessly integrated with various negative sampling methods and optimized using diverse solvers. To further capitalize on geographic information, we introduce proximity-aware negative samplers. This approach involves the probabilistic selection of negative locations, giving a higher probability to more informative negative locations. 

Overal, the contributions of our work can be succinctly summarized as follows:
\begin{itemize}
\item We introduce the Proximity-aware based region representation for sequential recommendation (PASR) built upon the Self-Attention Network, enhancing location recommendation. PASR effectively captures long-term sequential dependencies and optimally employs geographical data.
\item We propose a novel self-attention based geography encoder and a grid mapper. The encoder and mapper accurately represent the GPS coordinates of locations, enabling the capture of spatial proximity between nearby places. This approach models spatial clustering and distance-based location transitions better.
\item We utilize the novel loss function that employs importance sampling to optimize PASR. This approach assigns higher weights to informative negative samples, which in turn accelerates training. We also utilize proximity-aware negative samplers to enhance the quality of negative samples.
\item Our proposed PASR is rigorously evaluated using three real-world Location-Based Social Networking (LBSN) datasets. 
\end{itemize}

\section{Related Work}
In this section, We firstly concentrate on the advances of sequential location recommendation and then we illustrate the domain of sequential recommendation, especially the self-attention based solutions.

\subsection{Sequential Location Recommendation}
Sequential location recommendation has been approached through various modeling techniques, which includes interaction-based matrix factorization methods \cite{cheng2013you,lian2013collaborative},  metric embedding techniques \cite{feng2015personalized}, word embedding methodologies \cite{feng2017poi2vec}, and recurrent neural networks (RNNs) equipped with attention mechanisms \cite{cui2019distance2pre,li2018next,liu2016predicting,wu2019graph,zhao2020go}. The inclusion of geographical information has been achieved through the incorporation of travel distance \cite{cui2019distance2pre,li2018next}, the implementation of location transitions specific to distance \cite{liu2016predicting}, the utilization of geography-conscious uniform sampling \cite{cheng2013you}, or the incorporation of geography-aware gating mechanisms within RNNs \cite{zhao2020go}. Temporal information have also been integrated through techniques such as embedding time intervals \cite{li2018next}, time-of-the-week embedding \cite{li2018next,yang2017neural}, and the regulation of information flow using interval-aware gating \cite{zhao2020go}. These models have been optimized using various methods, including the Bayesian Personalized Ranking (BPR) loss function \cite{cheng2013you,cui2019distance2pre,feng2015personalized,lian2020personalized,lian2013collaborative,liu2016predicting}, cross-entropy loss \cite{li2018next,yang2017neural,zhao2020go}, and hierarchical softmax \cite{feng2017poi2vec}. The key distinctions between our proposed algorithm and these existing approaches encompass the geographic modeling techniques, the novel weighted loss function, and the incorporation of a self-attention based network to capture long-range dependencies.

\subsection{Methods for modeling geographical information}
The LBSN datasets have revealed the existence of spatial clustering patterns, which can be understood through Tobler's First Law of Geography \cite{ye2011exploiting,chen2023neural}. These patterns are further characterized by the distribution of distances between visited locations, displaying adherence to a power-law distribution. To circumvent assumptions related to power-law distributions, kernel density estimation has been utilized to estimate distance distributions between pairs of locations \cite{zhang2013igslr}. Nevertheless, modeling distance distributions may not fully account for the multi-center nature of one's visited places \cite{cheng2012fused}, prompting the development of geo-clustering methods aimed at identifying and grouping such locations \cite{cheng2012fused,liu2013learning}. Estimating the number of clusters can be a challenging task, and as a solution, 2-D kernel density estimation has emerged to capture the spatial clustering phenomenon \cite{lian2018scalable,lian2014geomf,lian2018geomf++,zhang2014igeorec,LiuTL020}. These geographical modeling techniques have been incorporated into models in an ad-hoc manner.

Contrasting with these methods, we introduce a self-attention-based geography encoder and a grid mapping approach. The encoder and mapper can be integrated with the self-attention network, offering a unique approach to capture geographical factors.

\subsection{Self-attention based sequential recommendation}
 In this discussion, we will specifically focus on self-attention based sequential recommendation. For a broader array of sequential recommendation algorithms, you may find more information in the provided survey references \cite{quadrana2018sequence,wang2021survey}. The self-attention network, known for its complete parallelism and the capability to capture long-range dependencies, has been extensively employed in sequence modeling. It has delivered state-of-the-art results across various domains, including Natural Language Processing (NLP) \cite{vaswani2017attention,LiuLCT2022,liu2021effective}, combinatorial optimization \cite{ZhaoLYR21,LIUTY2021,LiuZTY23}, and social recommendation. In recent years, this architecture has also been harnessed for sequential recommendation tasks, optimizing performance using the binary cross-entropy(BCE) loss based on inner product preferences \cite{kang2018self} or the triplet margin loss based on Euclidean distance preferences \cite{zhang2018next}. Empirical results show its significant performance improvement over traditional RNN-based methods. Nevertheless, the original self-attention network, initially designed for symbolic sequences, lacks inherent consideration for varying time intervals between consecutive interactions when modeling sequential dependencies \cite{LiuT019,LiuTY22,TangLYY21,LiuP023}. To address this limitation, the self-attention based network with time interval awareness was introduced \cite{li2020time}, refining attention weights by incorporating time intervals. Moreover, to enhance recommendation performance while avoiding information leakage, an approach reminiscent of the Cloze task used in BERT training \cite{devlin2018bert} was adopted \cite{sun2019bert4rec}, replacing the causality mask.

PASR sets itself apart from these methodologies through its innovative geography encoder and novel loss function. Combining these approaches could potentially lead to even more enhancements in recommendation accuracy.

\section{Preliminary: Attention mechanism}
To capture the long-term dependency within sequences and better extract features from the sequence, we utilize the encoder based on self-attention mechanism \cite{kang2018self} as our sequence encoder to transform the input embedding matrix $\mathbf{E}\in \mathbb{R}^{m\times d}$ which maps a sequence with length $m$ into $d$ dimentional space. Concretely, the self-attention encoder is constructed by stacking multiple self-attention modules, with each module consisting of a self-attention layer (SA) and a feed-forward network layer (FFN).

The self-attention layer computes new representations for each element by considering relationships with all other elements in the sequence. In the self-attention layer, the input is taken from either the embedding matrix $\mathbf{E}$ or the output of the previous self-attention module (referred to as $\mathbf{E}$). This input is then transformed using three separate linear projection matrices: $\mathbf{W}_Q, \mathbf{W}_K, \mathbf{W}_V \in \mathbb{R}^{d\times d}$. These transformations are then followed by an attention mechanism to obtain the final output:
$$
\mathbf{S} = SA(\mathbf{E}) = Attention(\mathbf{E}\mathbf{W}_Q, \mathbf{E}\mathbf{W}_K, \mathbf{E}\mathbf{W}_V)
$$
where the attention is the scaled dot-product attention, i.e., 
$$
Attention(\mathbf{Q}, \mathbf{K}, \mathbf{V})= softmax(\frac{\mathbf{Q}\mathbf{K}^T}{\sqrt{d}})\mathbf{V}
$$
In this attention,  $\mathbf{Q}$ represents the query,  while $K$ and $V$ represent the key and value, respectively. Intuitively, it calculates a weighted sum over all values, where the weights are given by the dot product between the query and the key. The scaling factor $\sqrt d$ is used to prevent the resulting dot product from becoming too large, which can lead to extreme weights, especially when the vector dimension $d$ is large.

It's important to highlight that the prediction for the $n+1$-th location relies solely on the preceding n behaviors, not future ones. To ensure adherence to this causality constraint, we enforce it through the use of a square mask. This mask is constructed with  $\text{-}\infty$ in its upper triangle and $0$ in others. This approach effectively prevents information leakage from future behaviors into the prediction process.

While self-attention can adaptively gather information from a user's historical behavior sequence, it is fundamentally a linear model. To introduce non-linearity to the model and consider interactions between different dimensions, we have incorporated the following feed-forward network layer into the self-attention module:
$$
\mathbf{F} = FFN(\mathbf{S})= \max(0, \mathbf{S}\mathbf{W}_1 + \mathbf{b}_1)\mathbf{W}_2 + \mathbf{b}_2
$$
where $\mathbf{W}_1 \in \mathbb{R}^{d\times d_h}$, $\mathbf{W}_2\in \mathbb{R}^{d_h\times d}$, $\mathbf{b}_1\in \mathbb{R}^{d_h}$, $\mathbf{b}_2\in \mathbb{R}^d$, and they satisfy the condition $d_h > d$.

To capture more intricate features within the sequence, we stack $N$ self-attention modules. However, as the neural network becomes deeper, issues such as vanishing gradients and slow training speed can arise. To ensure a stable training process and expedite training speed, we employ residual connections and layer normalization \cite{vaswani2017attention} on each layer of the network. This leads to the formation of the final sequence encoder:
$$
\mathbf{S}^l = \mathbf{F}^{l-1} + LayerNorm(SA(\mathbf{F}^{l-1}))\\
\mathbf{F}^l = \mathbf{S}^l + LayerNorm(FFN(\mathbf{S}^l))
$$
where $\mathbf{F}^0 = \mathbf{E}$ stands for the embedding matrix that acts as the input to the encoder, $\mathbf{S}^l$ refers to the output of the $l$-th layer of self-attention,  $\mathbf{F}^l$ represents the output of the $l$-th layer of the feed-forward neural network, and the index $l\in \{1,2,\cdots,N\}$.

In general, the input of the self-attention based module is a sequence represented by a matrix $\mathbf{E}\in \mathbb{R}^{m\times d}$ and the outpu is also a matrix with the same shape as $\mathbb{E}$. The main effect of self-attention mechanism is to conduct the intra-interaction with the sequence. In the following sections, the self-mechanism based module is used in the Geography Encoder and self-Attention Encoder.

\section{PASR method}

Let $L=\{l_1, l_2, \cdots, l_Q\}$ represents a set of $Q$ locations, $U=\{u_1, u_2, \cdots, u_M\}$ represents a set of $M$ users, and $S=\{s^{u_1}, s^{u_2}, \cdots, s^{u_M}\}$ represents a set of historical mobility sequences for all users. A sequential location recommender operates on a user's mobility trajectory, which is denoted as $S^u= r_1^u \rightarrow r_2^u \rightarrow \cdots \rightarrow r_n^u$.  In this representation, $r_i^u = (u, l_i, \alpha_i, \beta_i)$ signifies a user behavior, where $u$ indicates the user,  $l_i$ denotes the visited location, and  $\alpha_i$ and $\beta_i$ represent the latitude and longitude of that location, respectively. Given this input trajectory, the objective of a sequential location recommender is to predict the subsequent location $l_{i+1}$ along with its GPS position $p_{i+1}$. During the training phase, the model utilizes the user's trajectory excluding the final behavior $r^u_1 \rightarrow r_2^u \rightarrow \cdots \rightarrow r^{u}_{n-1}$ as the input sequence. And the output sequence encompasses the trajectory excluding the initial behavior $r_2^u \rightarrow r_3^u \rightarrow \cdots \rightarrow r_n^u$. The representation module of the Point-of-Interest (POI) is depicted in \textbf{Figure} \ref{fig_embedding_framework_pasa} and the overall framework of the PASA is depicted in \textbf{Figure} \ref{fig_overall_framework_pasa}. Each component of this architecture will be elaborated upon in the subsequent sections.

\subsection{Embedding Representation Module for POI}
\begin{figure}[t]
\centering
\includegraphics[width=13.5cm]{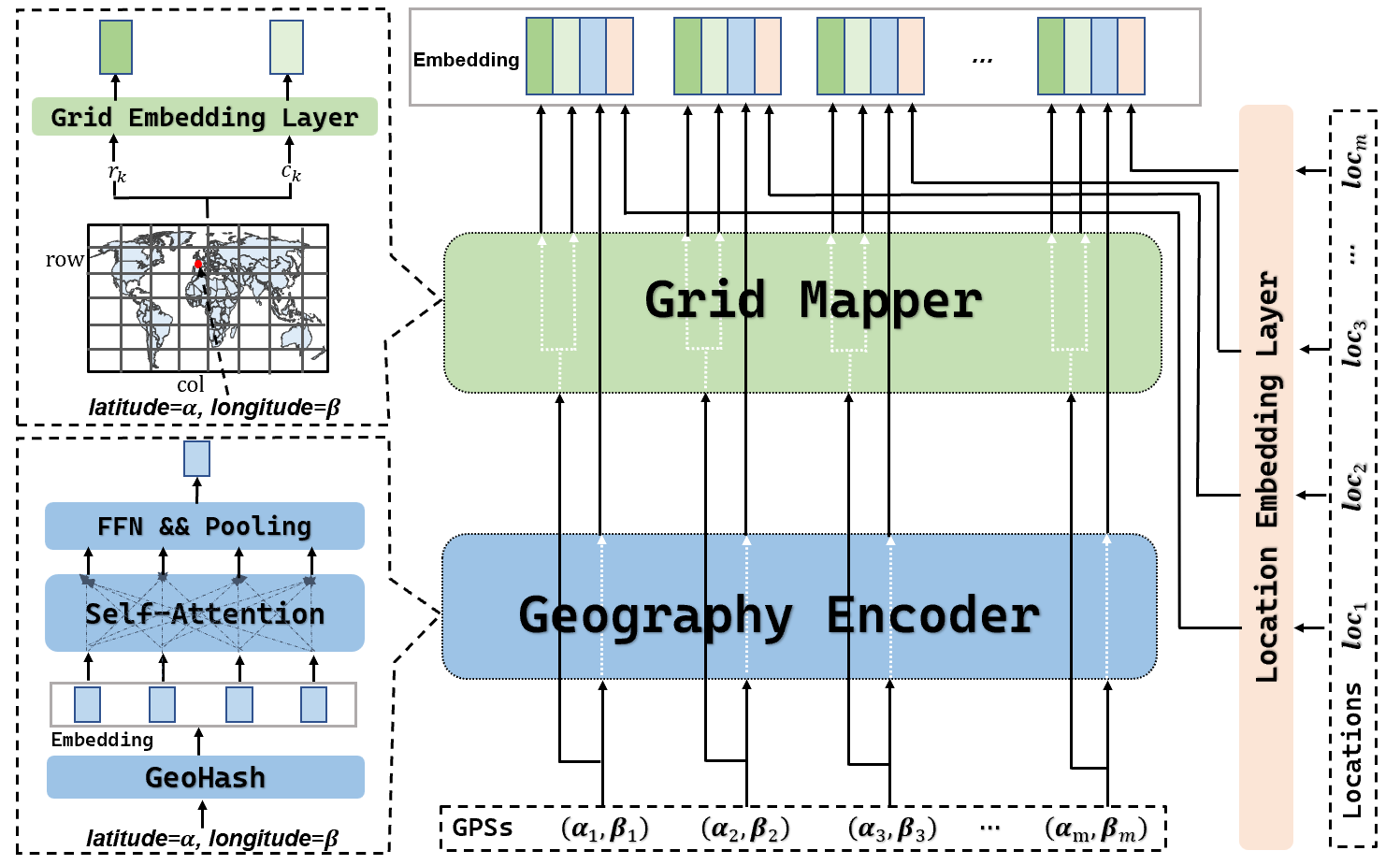}
\caption{The embedding representation module of Point-of-Interest}
\label{fig_embedding_framework_pasa}
\end{figure}

For user $u$, we can extract her/his interests from the historical mobility sequence $S^u= r_1^u \rightarrow r_2^u \rightarrow \cdots \rightarrow r_n^u$.  For the ease of parallelism, we preprocess the input sequence $r^u_1\rightarrow r^u_2 \rightarrow \cdots \rightarrow r^u_{n}$ into a sequence of fixed length $m$. If the length of the input sequence exceeds the given value $m$, we divide it into multiple sub-sequences, each with a length of $m$. In cases where the input sequence is shorter than $m$, we apply a "padding" operation at its right end, extending it until it reaches a length of $m$. To better represent the mobility sequence, three paradigms are utilized to embed the sequence, i.e. location ID embedding (w.r.t. Location Embedding Layer), GeoHash Embeding (w.r.t. Geography Encoder) and the Proximity grid embedding (w.r.t. Grid Mapper) as depicted in \textbf{Figure} \ref{fig_embedding_framework_pasa}.

\paragraph{$Location\  Embedding\  Layer$:}
Each location $l\in \{l_1, l_2, \cdots, l_Q\}$ corresponds to an one-hot embedding with dimension $d$, which comprises the location embedding matrix $\mathbf{M}_{loc}\in \mathbb{R}^{Q\times d}$. So that the historical mobility sequence with length $m$ can be represented by a matrix $\mathbf{E}_{loc}\in \mathbb{R}^{m\times d}$.

\paragraph{$Geography\ Encoder\ Layer$:}
The geographical information of a location is typically represented by its longitude and latitude coordinates. While it's theoretically possible to directly feed these floating-point values into the learning system, it's infeasible due to two reasons. First, as human activities involve a tiny fraction of the entire Earth's surface, using raw latitude and longitude values directly leads to severe sparsity problems. Second, latitude and longitude are closely intertwined because both are needed together to precisely determine a location. The complex interaction between these two attributes could pose a challenge for the learning system to fully comprehend and utilize. To tackle these challenges, we utilize the geography encoder a solution. 

For any input geography coordinate, the geography encoder first utilizes the Geohash to dispose of it. Geohash is a public-domain geocode system, which encodes a geography coordinate into a string of letters and digits. It is a hierarchical spatial data structure that subdivides space into buckets of grid shape. Specifically,  the entire map is first divided into $32$ grids, each sub-grid is assigned a number or letter as its identifier using the Base$32$ encoding scheme. Then, this process is applied recursively to the sub-grids until a designated precision level $L$ is attained. At this stage, the entire map is divided into $32^L$ small grids, and each small grid is assigned a unique encoding string of length $N$ composed of Base$32$ characters. The geographical coordinates of all locations within a grid can be denoted by the corresponding encoding string of that grid.  For example, the GPS coordinates (40.68925, -74.0445) of the Statue of Liberty are converted into a geohash string $'dr5r7p62n1'$ through the Geohash system (seeing \textbf{Figure} \ref{fig_geohash}), with a string length specified as 12. In this manner, the encoding process leverages both latitude and longitude,  effectively handling the inherent interaction between them. Moreover, different locations in the same region will share the same geohash string, mitigating the sparsity issue to a certain extent.

It's straightforward and intuitive to embed the encoding strings with an embedding matrix, but the number of encoding strings are exponential which may suffer from serious issue of sparsity, and the proximity relationships between the nearby locations cannot be captured effectively.  To this end, we view an encoding string as a string of single characters, where each character belongs to the set of Base32 characters. However, it's important to note that a character-level model is unable to fully capture the proximity relationship between nearby locations \cite{abs-2302-02568}. To address this limitation, we first transform each string into a sequence of n-grams, thereby expanding the vocabulary size from $32$ to $32^n$. For instance, let's consider the initial 6 characters-$'dr5r7p'$ of the aforementioned string. It is transformed into a sequence of bigrams:  $dr \rightarrow r5 \rightarrow 5r \rightarrow r7 \rightarrow 7p$. Once we obtain the n-gram sequence, we embed this sequence and subsequently apply a stacked self-attention based network to capture sequential dependencies. Following this, we aggregate the representations of the n-gram sequence using an average pooling technique, seeing the left-down subfigure of \textbf{Figure} \ref{fig_embedding_framework_pasa}. By this way, each geography coordinate of location in the historical mobility sequence can be embeded into a vector with $d$ dimension by Geohash encoding and Goegraph Encoder.

\paragraph{$Grid\ Mapper$:}
After applying the geography encoder to the GPS coordinate, we have obtained a good representation of geography information of the location. But note that the using of Geohash still exists some limitations.  First, the Geohash system could be faced with edge cases, where two close locations have completely different geohash strings. For example, the distance between GPS coordinates $(-0.005, 90)$ and $(0.011, 90)$ is only about $1.78$ km, but their corresponding encoding string is $'qpbpbp04b5bj'$ and $'w00004000481'$, respectively. Second, although the use of n-gram in transforming the encoding string could improve the quality of the representation of geography information, the larger vocabulary size may exacerbate the sparsity problem. To solve the challenges mentioned above, we propose using the grid mapper to reprocess the latitude and longitude, which will make better use of the geography information and can be viewed as a supplementary of the geography encoder.

The grid mapper only considers the valid region which contains locations and then divides the map into grids based on latitude and longitude. Concretely, we first traverse the GPS coordinates of all locations and record the maximum and minimum values of latitude and longitude respectively during pre-processing. In the subsequent steps, we only focus on the region bounded by these recorded maximum and minimum values of latitude and longitude, which helps alleviate sparsity issues. We treat this region as a flat surface and divide it into grids along both latitude and longitude. Latitude corresponds to rows, while longitude corresponds to columns. With this setup, the region is divided into numerous grids, and each grid is indexed by the row ID and column ID (seeing \textbf{Figure} \ref{fig_grid_sketch}). Any given location falls within one of these grids, and its GPS coordinates are represented by the ID pairs of the corresponding grid. Following this, we directly embed the row ID and column ID into dense vectors with dimension $d$, seeing the right-top subfigure of \textbf{Figure} \ref{fig_embedding_framework_pasa}.

\paragraph{Representation of the historical mobility sequence:}
As the aforementioned description, We create a location embedding matrix $\mathbf{M}_{loc} \in \mathbb{R}^{Q\times d}$, where $d$ is the dimension of the feature vectors, to convert one-hot encodings of interest points into dense embedding vectors. Thus, after applying the $\mathbf{M}_{loc}$ matrix transformation to the input sequence of length $m$, we obtain an embedding matrix $\mathbf{E}_{loc} \in \mathbb{R}^{m\times d}$ to serve as subsequent input. For GPS sequences with length $m$ composed of latitude and longitude, we utilize a  geographic encoder and a grid mapper to transform them into a geographic embedding matrix  $\mathbf{E}_{geo}\in \mathbb{R}^{m\times d}$,  a row embedding matrix $\mathbf{E}_{row}\in \mathbb{R}^{m\times d}$ , and a column embedding matrix $\mathbf{E}_{col}\in \mathbb{R}^{m \times d}$.  We now have obtained the location embedding matrix $\mathbf{E}_{loc}$, the geographic embedding matrix $\mathbf{E}_{geo}$, the row embedding matrix $\mathbf{E}_{row}$, and the column embedding matrix $\mathbf{E}_{col}$. These four matrices are concatenated to form the embedding matrix $\mathbf{E}_{input}\in \mathbb{R}^{m\times 4d}$, which serves as input to the following network. 
As we use a sequence encoder based on the self-attention mechanism to transform the input matrix, it differs from recurrent or convolutional modules and cannot capture positional information in the sequence. Similar to the approach in \cite{kang2018self}, we add a learnable positional embedding matrix $\mathbf{P} \in \mathbb{R}^{m\times 4d}$ to the input embedding matrix $\mathbf{E}_{input}$, that is, $\mathbf{E}_{input} = \mathbf{E}_{input} + \mathbf{P}$.

\subsection{Attention-based Decoder}

\begin{figure}[t]
\centering
\includegraphics[width=13.5cm]{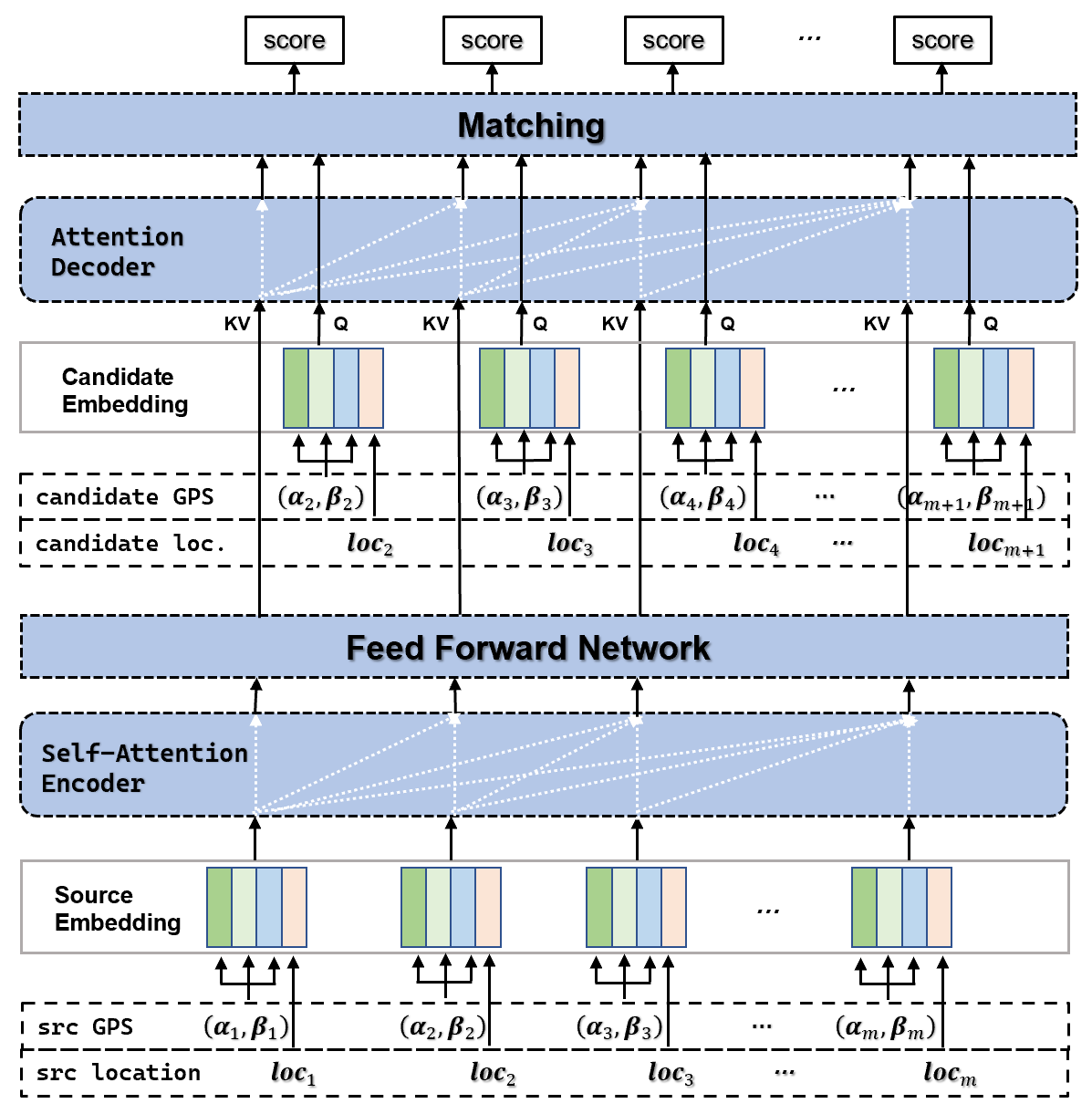}
\caption{The overall framework of PASA}
\label{fig_overall_framework_pasa}
\end{figure}

The over framework of PASR is illustrated in \textbf{Figure} \ref{fig_overall_framework_pasa}. For a user, the corresponding historical mobility sequence is feed into the embedding representation module (seeing in \textbf{Figure} \ref{fig_embedding_framework_pasa}) to obtain the source embedding. Then the source embedding is feed into the self-attention based encoder followed by a feed forward network to get the hidden representation (denoted as $\mathbf{F}^N\in \mathbb{R}^{m\times 4d}$) of the historical mobility sequence. As we aim to predict the next location, the historical mobility sequence consists of the next locations (i.e. the sequence consists locations from second to $m+1$-th location) is regarded as the candidate target and the corresponding candidate embedding (denoted as $\mathbf{T}\in \mathbb{R}^{m\times 4d}$) as derived according to the embedding representation module (a.k.a. \textbf{Figure} \ref{fig_embedding_framework_pasa}).

Many of the current recommendation systems based on self-attention mechanism directly pass the encoder's outputs to the matching module. However, recent research findings \cite{lian2018xdeepfm,zhou2018deep} suggest that this approach may not be optimal. In order to improve the representation of the input sequence with respect to target locations, PASR introduces a decoder with target-aware attention. The mechanism of this decoder is as follows:
$$
\mathbf{A}= decoder(\mathbf{F}^N|\mathbf{T})=Attention(\mathbf{T}, \mathbf{F}^N\mathbf{W}, \mathbf{F}^N)
$$
where $\mathbf{T}\in R^{m\times 4d}$ is the representation matrix of the output sequence and $\mathbf{W}\in R^{4d\times 4d}$ serves the purpose of projecting queries and keys into a common latent space. Importantly, it should be noted that the causality constraint remains a necessity, which is accomplished through the utilization of the mask as previously mentioned.
\subsubsection{$Prediction\ scores$}
When provided with the representation $\mathbf{A}_i$ of the input sequence at the step $i$, the preference scores for candidate locations can be computed using any matching function $f$. This matching function could be a deep neural network \cite{zhou2018deep}, the dot product \cite{kang2018self} when the historical and candidate representations have identical dimensions, or a bilinear function when their dimensions differ. Concretely, preference scores are formulated as follows:
$$
y_{i,j}=f(\mathbf{A}_i, \mathbf{T}_j)
$$
where $T_j$ represents the feature vector for candidate location $j$, derived by concatenating the embedding vectors of that location and the feature vectors of its geographical information. Just like in \cite{kang2018self}, the embedding matrices, the geography encoder, and the grip mapper are common to both the input sequences and the output sequences.

\subsection{Importance sampling based loss}
Given the sequence $S^u$, the candidate location $j$'s preference score at step $i$ is denoted as $y_{i,j}$. It's evident that optimizing the commonly used cross-entropy loss \cite{li2018next,yang2017neural,zhou2018deep} is inefficient in such scenario when the number of candidate locations becomes substantial. In scenarios where self-attention based sequential recommenders are concerned, the binary cross-entropy(BCE) loss is commonly adopted \cite{kang2018self,li2020time}. This loss function can be represented as follows:
$$
\mathcal{L}= -\sum_{S^u\in S}\sum_{i=1}^{n}\left(\log\sigma\left(y_{i,t_i}\right)+\sum_{l \notin L^u}\log\left(1-\sigma\left(y_{i,l}\right)\right) \right)
$$
where $S$ represents the set consisting of users' mobility trajectories, $L^u$ denotes the set of locations visited by user u, and $t_i$ signifies the target location at time step $i$. In this context, we have already excluded the padding item from the loss calculation. For effective loss optimization, one negative location is sampled from user $u$'s unvisited locations at each time step through a uniform distribution.

As only one negative location is randomly selected from user's unvisited locations, the BCE loss may not efficiently leverage the abundance of unvisited locations.  Specifically, once the loss has been optimized for a few epochs, positive locations become readily distinguishable from randomly chosen negative samples. Consequently, the gradient of the loss decreases dramatically, leading to sluggish training progress. Essentially, the absence of informative negative samples obstructs the optimization of the BCE loss. Intuitively, unvisited locations with high preference scores should  exert a more substantial influence on the gradient. These locations inherently possess more valuable information and should be sampled with higher probabilities. Nevertheless, directly selecting the top-$k$ unvisited locations with the greatest scores as negatives isn't feasible due to the risk of introducing false negatives. And directly sampling negative locations in proportion to their preference scores is also infeasible due to the efficiency challenges. To address these challenge, We adopt the similar approach proposed in the literature \cite{lian2020geography} to assign weights to unvisited locations based on negative probabilities. By employing this method, we ensure that even when using a uniform sampler, locations with more information can receive enhanced attention. Specifically, the BCE loss function is restructured as follows:
$$
\mathcal{L}= -\sum_{S^u\in S}\sum_{i=1}^{n}\left(\log\sigma\left(y_{i,t_i}\right)+\sum_{l \notin L^u}P\left(l|i\right)\log\left(1-\sigma\left(y_{i,l}\right)\right) \right)
$$
where $P(l|i)$ represents the probability of the location $l$ being negative given the user $u$'s mobility trajectory $r_1^u \rightarrow r_2^u \rightarrow \cdots \rightarrow r_i^u$. We suggest modeling the probability being negative as follows
$$
P(l|i) = \frac{\exp\left(y_{i,l}/T\right)}{\sum_{l'\notin L^u}\exp(y_{i,l'}/T)}
$$
where $T$ represents a temperature parameter that controls the deviation of the proposal distribution from a uniform distribution. As $T$ tends towards infinity, the distribution approaches uniformity.

Despite this, the efficiency of the restructured loss function is still hampered by the computational burden of normalizing probabilities. To enhance efficiency, while accounting for $\sum_{l \notin L^u} P(l|i)\log(1-\sigma(y_{i,l}))$, we adopt the notion of expectation computation in relation to $P(l|i)$. we introduce an approach that approximates this expectation through importance sampling. Let's assume a proposal distribution denoted as $Q(l|i)$ from which simple sampling can be carried out. Furthermore, we represent $\tilde{Q}(l|i) $ as the unnormalized probability of $Q(l|i)$. With inspiration drawn from \cite{10.5555/1162264}, the loss is then approximated in the subsequent manner:
$$
\mathcal{L}= -\sum_{S^u\in S}\sum_{i=1}^{n}\left(\log\sigma\left(y_{i,t_i}\right)+\sum_{l \notin L^u}w_l\log\left(1-\sigma\left(y_{i,l}\right)\right) \right)
$$
where $w_l = \frac{\exp(y_{i,l}/T-\ln \tilde{Q}(l|i))}{\sum_{l'=1}^l\exp(r_{i,l'}/T-\ln \tilde{Q}(l'|i))}$ is the weight of the $l$-th sample. Consequently, within the set of $l$ locations, those with higher preference scores are allocated greater weights.  In scenarios where the proposal distribution $Q$ departs from distribution $P$, the weight serves to mitigate the deviation between $P$ and $Q$ to a certain degree.

It's important to highlight that we exclusively make use of the unnormalized probability, a choice that proves advantageous when working with probability distributions that pertain to a subset of the overall location set $L$. In cases where the proposal distribution $Q(l|i)$ assumes a uniform distribution over $L\backslash L^u$, we observe that $\ln \tilde{Q} (l|i) \propto -\ln |L|$, which results in the weight $w_l = \frac{\exp(y_{i,l})/T}{\sum_{l'=1}^K\exp(y_{i,l'}/T)}$. Approximating $Q(l|i)$ with the uniform distribution over $L$ is a valid approach, primarily due to the minimal probability associated with sampling locations from $L^u$ as negative samples and the comparable level of recommendation accuracy. When devising alternative proposal samplers, we adopt a similar strategy, considering the distribution over $L$ instead of $L\backslash L^u$ to simplify the process. This method is widely employed in the field of Natural Language Processing (NLP) \cite{mikolov2013efficient}.

\subsection{Proximity-aware negative sampler}

For a sequential location recommender, the geographical information can also play a pivotal role in distinguishing between negative and potentially positive ones within unvisited locations. For instance, when a user visits the location $t$, the unvisited locations proximate to $t$ may be more inclined to be negative. Nevertheless, directly sampling locations grounded in GPS distance might be computationally  infeasible. In response to this challenge, the proximity-aware negative samplers adopt a pragmatic approach. We first retrieve the initial retrieval of  $K$  nearest locations to the target location. Subsequently, negative samples are randomly drawn from these $K$ candidates. This negative sampling procedure can be based on either a uniform distribution or a distribution based on location popularity. In the context of the popularity-based proposal distribution, our empirical findings underscore the efficacy of utilizing $\tilde{Q}(l|i)\propto \ln(c_l+1)$,  where $c_l$ represents the occurrence frequency in the mobility history. Our proposed distribution formulation can effectively guide the negative sampling process.

\section{Experiment}

\subsection{Datasets}
We evaluate our method using three publicly available LBSN (Location-Based Social Networking) datasets: Gowalla, Brightkite, and Weeplaces. These datasets are commonly used geographical location datasets that offer a substantial amount of user location and behavior data, making them highly valuable for applications such as location-based recommendations and urban planning. Below is a brief description of these three datasets and their sources:

Gowalla dataset is provided by the Gowalla social networking platform, which operated from 2007 to 2012. The Gowalla dataset includes information such as the geographic locations of user check-ins, timestamps, and the points of interest they checked into. It is a large-scale geographical location dataset, comprising millions of check-in records and hundreds of thousands of points of interest.

Brightkite was a location-based social media network launched in 2007 and disbanded in 2012. The platform allowed users to check in at locations after visiting them through text messages or mobile applications.  It is a moderately large geographical location dataset, containing millions of check-in records and tens of thousands of points of interest.

Weeplaces is a service that visualizes users' check-in behavior on maps and has been integrated into several LBSN APIs. Users can log in to Weeplaces using their accounts from LBSNs and connect with friends who have also used the application on the same LBSN.  It is a medium-sized dataset comprising millions of check-in records and tens of thousands of points of interest.

Similar to the approach in \cite{lian2014analyzing}, we conducted data filtering by removing users with fewer than 20 check-ins and locations with fewer than 10 visits. Table \ref{tab:statics} summarizes the statistical information for the three datasets after this filtering process. For each user, we constructed a check-in sequence based on their check-in records in chronological order. For each check-in sequence, we selected the user's last check-in at an unvisited point of interest before that moment for testing and used the check-in records before that for training. This ensures that the evaluation is focused on predicting locations that the user has not visited in their historical records, making it more consistent with real-world scenarios. The maximum sequence length was set to 50. If a sequence exceeded 50, we divided it from right to left into non-overlapping sub-sequences of length 50 and used the most recent 50 check-in records for testing.

\begin{table}[h]
    \centering
    \caption{\textbf{Statistics of datasets.}}
    \begin{tabular}{c l l l}
        \toprule
         ~ & Gowalla& Brightkite& Weeplaces\\  
        \midrule
         \#locations&  $131,327$&	$48,177$&	$127,859$ \\
         \midrule
         \#users&  $31,708$&	$5,186$& $13,801$ \\
         \midrule
         \#check-ins& $2,963,324$&	$1,661,161$&	$5,328,782$\\
         \bottomrule
    \end{tabular}
    \label{tab:statics}
\end{table}
 
\subsection{Baselines}

To demonstrate the effectiveness of our proposed method, we compared it with the following benchmark models:

\textbf{BPR}\cite{rendle2012bpr}: A recommendation algorithm based on Bayesian personalized ranking, primarily used for handling implicit feedback data. The objective of the BPR algorithm is to learn a ranking function for each user that places items they prefer above items they dislike. BPR optimizes its loss function by maximizing posterior probabilities, resulting in low-dimensional latent vectors for users and points of interest (POIs). These latent vectors are used to calculate user preference scores for POIs and make recommendations.

\textbf{FPMC}\cite{cheng2013you}: A conventional recommendation system algorithm that combines ideas from matrix factorization and Markov chains to make personalized recommendations. In FPMC, the recommendation process is modeled as a sequence of events or transitions between items, and it leverages users' historical interactions to make recommendations.

\textbf{GRU4Rec}\cite{hidasi2015session}: A variant of recurrent neural networks  commonly used for processing sequential data. In the context of POI recommendation, GRU can be applied to model users' historical visit sequences to capture user interest evolution and temporal dependencies. By inputting a user's historical visit sequence into GRU, the model can learn similarities between users and associations between POIs, enabling it to make recommendations.

\textbf{SASRec}\cite{kang2018self}: A sequence recommendation model based on self-attention mechanisms, particularly suitable for handling user sequential behavior data. It encodes a user's historical check-in sequence into vector representations and employs self-attention mechanisms to model the relationships between different items in the sequence. This allows the model to consider both previously visited POIs and the current context information during recommendation, resulting in personalized recommendations.

\textbf{GeoSAN}\cite{lian2020geography}: A sequence recommendation also use self-attention mechanisms and incorporate geography information in the model.  It encodes a GPS coordinate to a quadkey string and then used a self-attention based encoder to convert the quadkey string to an embedding with geographical information. The representation of one POI is the concatenation of  its id embedding and the geographical embedding.  The major difference between GeoSAN and our proposed model is the using of geohash algorithm and the introduction of grid mapper.

\subsection{Metric}
The performance of recommendations is evaluated based on the ranking of interest points that users actually visited within the recommended list provided by the recommendation system. We assess our POI recommendation model using two widely used evaluation metrics in recommendation systems: Hit Rate (HR@k) and Normalized Discounted Cumulative Gain (NDCG@k):

\textbf{HR@k} (Hit Rate) refers to the proportion of users, whose true interest points that appear in the top k positions of the recommended list generated for the user. It emphasizes the accuracy of the model's recommendations, i.e., whether the items of interest to the user appear in the recommended list. Its formula is as follows:
$$
\mathrm{HR}@k= \sum\limits_{u\in U} \frac{I(R_u\cup T_u \neq\emptyset)}{|U|}
$$
where $U$ represents the set of users, $R_u$ is the recommended list generated for user $u$, $T_u$ is the true set of relevant items for user $u$, and $I(\cdot)$ is the indicator function, which equals 1 if the condition inside the parentheses is true, and 0 otherwise.

\textbf{NDCG@k} (Normalized Discounted Cumulative Gain) refers to the cumulative gain of relevant items in the top k positions of a recommended list, divided by the maximum possible cumulative gain under ideal conditions. It emphasizes the utility of the model's recommendations, i.e., whether relevant items are ranked reasonably in the recommended list. NDCG@k can handle differences in relevance between different items, giving higher weight to more relevant items. Its formula is as follows:
$$
\mathrm{NDCG}@k=\frac{\mathrm{IDCG}@k}{\mathrm{DCG}@k} 
$$
where $\mathrm{DCG}@k$ represents the Discounted Cumulative Gain for the top k positions in the recommended list, taking into account the positional influence. It is calculated as:
$$
\mathrm{DCG}@k=\sum\limits_{i=1}^k \frac{\mathrm{rel}_i}{\log_2(i+1)}
$$
here, $\mathrm{rel}_i$ is the relevance score of the recommended item at position $i$. In experiments, we set the relevance score for relevant locations as $\mathrm{rel}_i=1$ and for irrelevant locations as $\mathrm{rel}_i=0$. $\mathrm{IDCG}@k$ represents the Ideal $DCG@k$, which is the maximum possible $DCG@k$ for a recommended list sorted by relevance scores in descending order.

A set of 100 locations is randomly chosen to serve as negative candidates for ranking alongside the target location. Subsequently, we can calculate the Hit Rate and NDCG metrics based on the ranking of these 101 locations.  

\subsection{Setting}

We set the dimensions of interest point embedding vectors, geographic embedding vectors, grid row embedding vectors and grid column vectors as $d=50$, the latent space dimension $d_h$ as $128$, and the number of layers for the self-attention layers in the sequence encoder, geographic encoder, and target-aware attention decoder as $2$. The the number of interval in grid mapper is set to $5000$, i.e., we divide the valid region into $5000 * 5000$ grids. And we use a $3$-gram tokens to represent the geohash string to utilize the geography information. For each location within the dataset, we first retrieve its nearest $K=2000$ locations for the proximity-aware negative sampler to sample from. The number of negative samples for training is set to 5 on all three datasets.  To train the proposed model, we use the Adam optimizer to optimize the loss function, with a learning rate of $0.001$ and weight decay of $0.0001$. We perform $20$ epochs of training on each dataset. For the benchmark models we are comparing against, in order to facilitate a fair comparison, we set their model parameters and training processes the same wherever applicable.

\subsection{Comparison with Baselines}

Table \ref{tal:result} presents the HR and NDCGmetrics for our model and benchmark models on three datasets. From the experimental results, it can be observed that the performance in point-of-interest (POI) recommendation tasks progressively improves from BPR, FPMC, GRU, SASRec, GeoSAN, to our proposed model.

BPR performs poorly in the POI recommendation task, likely due to its reliance on training solely based on user's implicit feedback, failing to capture sequential effects in user check-in behaviors. On the other hand, FPMC models the transition relationships effectively using Markov chains, thus significantly improving recommendation performance compared to BPR. GRU captures sequential effects in user check-in behaviors through its recurrent neural network structure, resulting in some performance improvement. SASRec effectively captures both short-term and long-term dependencies in sequences benefiting from self-attention mechanisms, leading to improved recommendation performance. GeoSAN, being a strong benchmark model, based on self-attention mechanisms and utilizing hierarchical gridding of GPS locations to effectively leverage geographic information, achieves superior recommendation performance. Comparing PASR to GeoSAN, we observe a $4.41\%$ to $8.46\%$ improvement in the HR@5 metric and a $2.76\%$ to $7.72\%$ improvement in the NDCG@5 metric. This demonstrates the effectiveness of PASR.

\begin{table}[]
\caption{\textbf{Comparison with baselines}}
\label{tal:result}
\begin{tabular}{ccrrrrrr}
\hline
Dataset                     & Metric  & \multicolumn{1}{c}{BPR}    & \multicolumn{1}{c}{FPMC}   & \multicolumn{1}{c}{GRU4Rec} & \multicolumn{1}{c}{SASRec} & \multicolumn{1}{c}{GeoSAN} & \multicolumn{1}{c}{PASR}            \\ \hline
\multirow{4}{*}{Gowalla}    & HR@5    & \multicolumn{1}{c}{0.0872} & \multicolumn{1}{c}{0.1783} & \multicolumn{1}{c}{0.1830}   & \multicolumn{1}{c}{0.2768} & \multicolumn{1}{c}{0.3311} & \multicolumn{1}{c}{\textbf{0.3591}} \\
                            & NDCG@5  & \multicolumn{1}{c}{0.0523} & \multicolumn{1}{c}{0.1058} & \multicolumn{1}{c}{0.1082}  & \multicolumn{1}{c}{0.1977} & \multicolumn{1}{c}{0.2378} & \multicolumn{1}{c}{\textbf{0.2522}} \\
                            & HR@10   & \multicolumn{1}{c}{0.1618} & \multicolumn{1}{c}{0.3471} & \multicolumn{1}{c}{0.3536}  & \multicolumn{1}{c}{0.3887} & \multicolumn{1}{c}{0.4610} & \multicolumn{1}{c}{\textbf{0.5096}} \\
                            & NDCG@10 & \multicolumn{1}{c}{0.0762} & \multicolumn{1}{c}{0.1597} & \multicolumn{1}{c}{0.1627}  & \multicolumn{1}{c}{0.2336} & \multicolumn{1}{c}{0.2795} & \multicolumn{1}{c}{\textbf{0.2969}} \\ \hline
\multirow{4}{*}{Brightkite} & HR@5    & 0.0850                     & 0.1873  & \multicolumn{1}{c}{0.1908}  & \multicolumn{1}{c}{0.3105}  & \multicolumn{1}{c}{0.4059 }                    & \textbf{0.4238}                     \\
                            & NDCG@5  & 0.0521                     & 0.1119  & \multicolumn{1}{c}{0.1144}  & \multicolumn{1}{c}{0.2027}  & \multicolumn{1}{c}{0.2897}                     & \textbf{0.2977}                     \\
                            & HR@10   & 0.1479                     & 0.3537  & \multicolumn{1}{c}{0.3634}  & \multicolumn{1}{c}{0.4285}  & \multicolumn{1}{c}{0.5607}                     & \textbf{0.5642}                     \\
                            & NDCG@10 & 0.0721                     & 0.1651  & \multicolumn{1}{c}{0.1695}  & \multicolumn{1}{c}{0.2374}  & \multicolumn{1}{c}{0.3398}                     & \textbf{0.3431}                     \\ \hline
\multirow{4}{*}{Weeplaces}  & HR@5    & 0.0812                     & 0.1778  & \multicolumn{1}{c}{0.1843}  & \multicolumn{1}{c}{0.2527}   & \multicolumn{1}{c}{0.3011}                     & \textbf{0.3207}                     \\
                            & NDCG@5  & 0.0482                     & 0.1055  & \multicolumn{1}{c}{0.1090}  & \multicolumn{1}{c}{0.1834}   & \multicolumn{1}{c}{0.2099}                     & \textbf{0.2261}                     \\
                            & HR@10   & 0.1526                     & 0.3458  & \multicolumn{1}{c}{0.3557}  & \multicolumn{1}{c}{0.3513}   & \multicolumn{1}{c}{0.4397}                     & \textbf{0.4555}                     \\
                            & NDCG@10 & 0.0710                     & 0.1592  & \multicolumn{1}{c}{0.1637}  & \multicolumn{1}{c}{0.2150}    & \multicolumn{1}{c}{0.2544}                     & \textbf{0.2694}                     \\ \hline
\end{tabular}
\end{table}


\subsection{Ablation Study}

To assess the impact of different elements within our approach, we carried out an ablation study. Our basic model (PASR) adopts the structure illustrated in \textbf{Figure} \ref{fig_overall_framework_pasa}   and takes into account the following model variants for comparison:
\begin{itemize}
    \item\textit{US (Uniform Sampler)}: Instead of using the proximity-aware negative sampler, a uniform sampler over all unvisited locations is employed during training.
    \item\textit{BCE (Binary Cross-Entropy) Loss}: We utilize the vanilla BCE loss without assigning weights to negative samples, essentially reverting to a simpler loss function.
    \item\textit{Remove GE (Geography Encoder)}: We remove the geography encoder from the model architecture, and the representation is constructed by concatenating the location embedding, grid row embedding, and grid column embedding.
    \item\textit{Remove GM (Grid Mapper)}: We remove the grid mapper and use the concatenation of location embedding and geographic embedding as the representation.
    \item\textit{Remove GE (Geography Encoder) and GM (Grid Mapper)}: Both the geography encoder and grid mapper are simultaneously removed, leaving only the location embedding as the representation.
    \item\textit{Remove AD (Attention-based Decoder)}: The attention-based decoder is removed from the model, and only the output of the encoder is utilized for matching.
    \item\textit{Add UE (User Embedding)}:  Each user is transformed into a dense embedding, which is then added to the location embedding.
    \item\textit{Add TE (Time Embedding)}:  Timestamps of check-in records are mapped to one-hour intervals within a week (resulting in 168 time intervals), and these intervals are fed into an embedding layer to obtain time embeddings.
\end{itemize}

The results of the ablation study are summarized in Table \ref{tal:ablation}. From the table, we can draw the following conclusions:
\begin{itemize}
    \item I. \textit{The proximity-aware negative sampler is helpful in certain circumstances.} Employing the proximity-aware negative sampler benefits the recommendation performance on the dataset Gowalla and Weeplaces, but leads a slight decline on dataset Brightkite. The reason behind that is maybe the number of locations in the Brightkite is relatively small, the using of proposed sampler may lead to sample some false negative samples.
    \item II. \textit{The proposed novel loss function demonstrates its efficacy.} Compared to the result where vanilla BCE loss is employed, our proposed loss can improve the performance by 1.69\%, 0.71\% and 2.61\% on the term of NDCG@5. This is because the proposed new loss can assign higher weight on the more informative negative samples to promote the training process.
    \item III. \textit{The utilizing of geography information plays a significant role to improve the recommendation performance.} We use geography encoder and grid mapper to utilize the geography information, and remove any part of them will lead to the degradation of recommendation accuracy. It's notable that the using of geography encoder and grid mapper can improve the performance dramatically, and the improvements are 8.61\%, 12.51\% and 10.45\% on the three datasets in terms of NDCG@5. It's interesting that only remove one of geography encoder and grid mapper will only lead a slight decline, which means that both the geography encoder and grid mapper can incorporate the geography information well.
    \item IV. \textit{Adding user embedding or time embedding plays no role on recommendation accuracy.} Adding time embedding only improves the NDCG@10 slightly on the Gowalla, while adding user embedding will cause a  dramatic performance degradation on the three datasets. This may because the introduction of time embedding or uesr embedding will lead to the mismatch between the candidate embedding space and the check-in embedding space.
    \item V. \textit{Using an attention-based decoder will improve the performance.} The improvements are 1.21\%, 1.65\% and 4.53\% on the three datasets in the term of HR@5.  The decoder makes the output of encoder to attend to historical check-ins relevant to the target, resulting in enhancements in recommendation performance.
    
\end{itemize}

\begin{table}[h]
\caption{\textbf{Ablation Study Results}}
\label{tal:ablation}
\resizebox{\textwidth}{!}{
\begin{tabular}{ccrrrrrrrrc}
\hline
Dataset                     & Metric  & \multicolumn{1}{c}{US}     & \multicolumn{1}{c}{BCE}    & \multicolumn{1}{c}{-GE}    & \multicolumn{1}{c}{-GM}    & \multicolumn{1}{c}{-GE-GM} & \multicolumn{1}{c}{-TAAD}  & \multicolumn{1}{c}{+UE}    & \multicolumn{1}{c}{+TE}             & \multicolumn{1}{c}{PASR}            \\ \hline
\multirow{4}{*}{Gowalla}    & HR@5    & \multicolumn{1}{c}{0.3494} & \multicolumn{1}{c}{0.3590} & \multicolumn{1}{c}{0.3554} & \multicolumn{1}{c}{0.3537} & \multicolumn{1}{c}{0.3328} & \multicolumn{1}{c}{0.3548} & \multicolumn{1}{c}{0.3313} & \multicolumn{1}{c}{0.3581}          & \multicolumn{1}{c}{\textbf{0.3591}} \\
                            & NDCG@5  & \multicolumn{1}{c}{0.2485} & \multicolumn{1}{c}{0.2480} & \multicolumn{1}{c}{0.2478} & \multicolumn{1}{c}{0.2474} & \multicolumn{1}{c}{0.2322} & \multicolumn{1}{c}{0.2520} & \multicolumn{1}{c}{0.2323} & \multicolumn{1}{c}{0.2505}          & \multicolumn{1}{c}{\textbf{0.2522}} \\
                            & HR@10   & \multicolumn{1}{c}{0.4827} & \multicolumn{1}{c}{0.5087} & \multicolumn{1}{c}{0.5026} & \multicolumn{1}{c}{0.4988} & \multicolumn{1}{c}{0.4684} & \multicolumn{1}{c}{0.4923} & \multicolumn{1}{c}{0.4699} & \multicolumn{1}{c}{0.5026}          & \multicolumn{1}{c}{\textbf{0.5096}} \\
                            & NDCG@10 & \multicolumn{1}{c}{0.2915} & \multicolumn{1}{c}{0.2962} & \multicolumn{1}{c}{0.2953} & \multicolumn{1}{c}{0.2941} & \multicolumn{1}{c}{0.2758} & \multicolumn{1}{c}{0.2964} & \multicolumn{1}{c}{0.2770} & \multicolumn{1}{c}{\textbf{0.2971}} & \multicolumn{1}{c}{0.2969}          \\ \hline
\multirow{4}{*}{Brightkite} & HR@5    & \multicolumn{1}{c}{\textbf{0.4288}}  & \multicolumn{1}{c}{0.4142}   & \multicolumn{1}{c}{0.4128}   & \multicolumn{1}{c}{0.4057}   & \multicolumn{1}{c}{0.3712} & \multicolumn{1}{c}{0.4169} & \multicolumn{1}{c}{0.3803} & \multicolumn{1}{c}{0.4136}  & \multicolumn{1}{c}{0.4238}          \\
                            & NDCG@5  & \multicolumn{1}{c}{\textbf{0.3049}}  & \multicolumn{1}{c}{0.2956}   & \multicolumn{1}{c}{0.2935}   & \multicolumn{1}{c}{0.2916}   & \multicolumn{1}{c}{0.2646} & \multicolumn{1}{c}{0.2952} & \multicolumn{1}{c}{0.2642} & \multicolumn{1}{c}{0.2929}  & \multicolumn{1}{c}{0.2977}          \\
                            & HR@10   & \multicolumn{1}{c}{0.5613}           & \multicolumn{1}{c}{0.5453}   & \multicolumn{1}{c}{0.5553}   & \multicolumn{1}{c}{0.5461}   & \multicolumn{1}{c}{0.4952} & \multicolumn{1}{c}{0.5565} & \multicolumn{1}{c}{0.5162} & \multicolumn{1}{c}{0.5561}  & \multicolumn{1}{c}{\textbf{0.5642}} \\
                            & NDCG@10 & \multicolumn{1}{c}{\textbf{0.3477}}  & \multicolumn{1}{c}{0.3410}   & \multicolumn{1}{c}{0.3395}   & \multicolumn{1}{c}{0.3366}   & \multicolumn{1}{c}{0.3047} & \multicolumn{1}{c}{0.3402} & \multicolumn{1}{c}{0.3080} & \multicolumn{1}{c}{0.3387}  & \multicolumn{1}{c}{0.3431}          \\ \hline
\multirow{4}{*}{Weeplaces}  & HR@5    & \multicolumn{1}{c}{0.3126}    & \multicolumn{1}{c}{0.3155}          & \multicolumn{1}{c}{0.3048}   & \multicolumn{1}{c}{0.2956}   & \multicolumn{1}{c}{0.2891} & \multicolumn{1}{c}{0.3068} & \multicolumn{1}{c}{0.2940} & \multicolumn{1}{c}{0.3121}  & \multicolumn{1}{c}{\textbf{0.3207}} \\
                            & NDCG@5  & \multicolumn{1}{c}{0.2196}    & \multicolumn{1}{c}{0.2202}          & \multicolumn{1}{c}{0.2151}   & \multicolumn{1}{c}{0.2073}   & \multicolumn{1}{c}{0.2047} & \multicolumn{1}{c}{0.2179} & \multicolumn{1}{c}{0.2055} & \multicolumn{1}{c}{0.2192}  & \multicolumn{1}{c}{\textbf{0.2261}} \\
                            & HR@10   & \multicolumn{1}{c}{0.4456}    & \multicolumn{1}{c}{0.4506}          & \multicolumn{1}{c}{0.4362}   & \multicolumn{1}{c}{0.4362}   & \multicolumn{1}{c}{0.4198} & \multicolumn{1}{c}{0.4412} & \multicolumn{1}{c}{0.4251} & \multicolumn{1}{c}{0.4544}  & \multicolumn{1}{c}{\textbf{0.4555}} \\
                            & NDCG@10 & \multicolumn{1}{c}{0.2624}    & \multicolumn{1}{c}{0.2668}          & \multicolumn{1}{c}{0.2574}   & \multicolumn{1}{c}{0.2526}   & \multicolumn{1}{c}{0.2467} & \multicolumn{1}{c}{0.2612} & \multicolumn{1}{c}{0.2476} & \multicolumn{1}{c}{0.2648}  & \multicolumn{1}{c}{\textbf{0.2694}} \\ \hline
\end{tabular}
}
\end{table}

\subsection{Performance w.r.t. Loss Function and Negative Sampling}
\begin{figure}
\centering
\includegraphics[width=12.5cm]{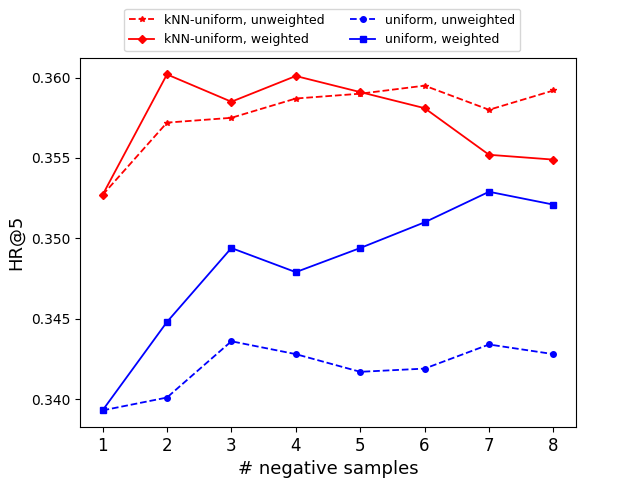}
\caption{The impact of using different loss functions, negative samplers and number of negative samples.}
\label{fig_negative_samples_hr_5.png}
\end{figure}

We investigate the effect of different combinations of loss function and negative samplers. Concretely, We experimented on the following four combinations: \textcircled{1}weighted loss(our proposed loss) + kNN-uniform sampler(our proposed sampler); \textcircled{2} unweighted loss(the vanilla BCE loss) + uniform sampler(sampling from all unvisited locations uniformly); \textcircled{3}unweighted loss + kNN-uniform sampler;  \textcircled{4}weighted loss + uniform sampler. For each combination of the loss function and the negative sampler, the number of negative samples is varied from 1 to 8.  The results on the Gowalla dataset are shown in \textbf{Figure} \ref{fig_negative_samples_hr_5.png}.

From the results, we can observe the following:
First, the model trained with the kNN-uniform sampler outperforms the model trained with the uniform sampler. This demonstrates the effectiveness of the proposed negative sampler. By sampling based on proximity, the kNN-sampler can select more informative negative instances, thereby enhancing the training process.Second, the weighted loss contributes significantly to model performance. The impact of the weighted loss is particularly pronounced when using the uniform sampler. When employing the kNN-sampler, the weighted loss facilitates performance improvement with a small number of negative samples. However, as the number of negative samples increases beyond a certain threshold, performance may experience a slight decline. This degradation could result from assigning higher weights to irrelevant or noisy negative samples as the number of negative samples increases. Therefore, selecting an appropriate number of samples is crucial in the training process.

\subsection{Sensitivity w.r.t Embedding Dimension}
\begin{figure}
\centering
\includegraphics[width=12.5cm]{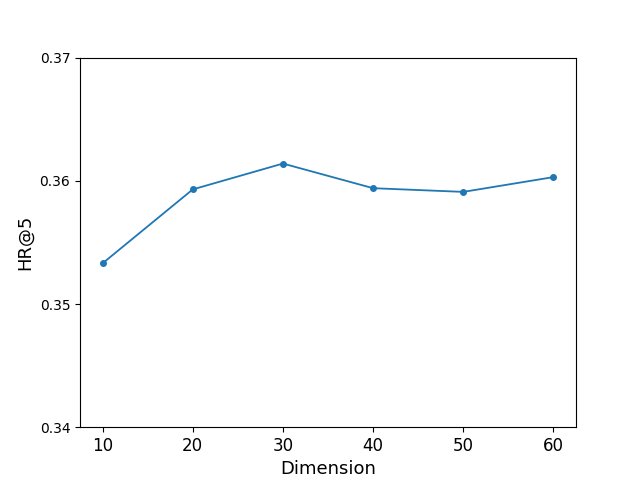}
\caption{The impact of embedding dimension}
\label{fig_embedding_dimension}
\end{figure}

The embedding dimension used in the model is varied from 10 to 60 with a step 10. \textbf{Figure} \ref{fig_embedding_dimension} shows the results. We observe a significant deterioration in performance when employing a small embedding dimension, whose expressive power is limited. Optimal recommendation accuracy is achieved with a moderate embedding size of 30, which can effectively capturing the inherent meaning of location and geographical data. Increasing the embedding dimension further, however, adversely impacts recommendation accuracy, as the number of locations and regions is limited.

\subsection{Sensitivity w.r.t Number of Interval in Grid Mapper}
\begin{figure}
\centering
\includegraphics[width=12cm]{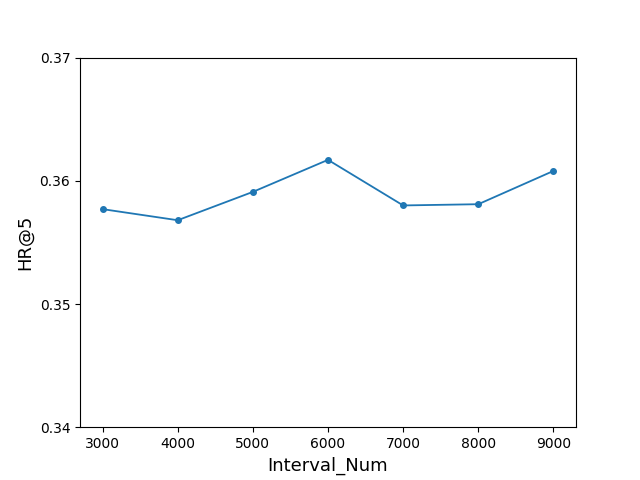}
\caption{The impact of number of interval in grid mapper}
\label{fig_interval_num}
\end{figure}

The number of interval used in the grid mapper is varied from 3000 to 8000 with a step 1000. \textbf{Figure} \ref{fig_interval_num} shows the results. The peak performance is obtained when the interval number is a medium value 6000. Too low or too high a number of interval can cause performance degradation. The former is due to lacking of differentiation between different regions, and the latter is because a large number of interval means the grid mapper divide the map into grate number of regions which could cause the sparsity problem. In general, the proposed model is insensitive to the interval number in the grid mapper.

\subsection{Sensitivity w.r.t N-gram}
\begin{figure}
\centering
\includegraphics[width=12.5cm]{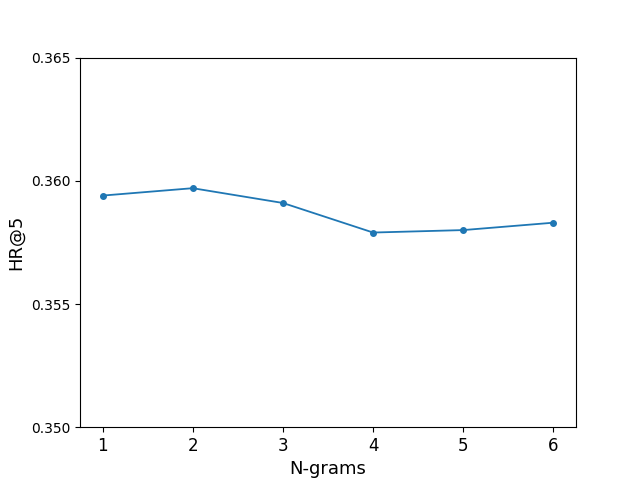}
\caption{The impact of N-gram}
\label{fig_n_gram}
\end{figure}

The $n$-grams used in the geography encoder is varied from $n = 1$ to $n = 6$. \textbf{Figure} \ref{fig_n_gram} show the results.
The peak performance is obtained when $n=2$. The size of gram token vocabulary is $32^n$, which will increase very rapidly as $n$ increases. The large $n$ will make the vocabulary size too large, and we can observe a slight decline of recommendation accuracy. In general, the proposed model is insensitive to $n$, and we recommend using a smaller $n$ for both better performance and smaller memory cost.

\vspace{3em}

\section{Conclusions}

In this paper, we propose a self-attention based method PASR for sequential location recommendation. To address the unbalance between positive samples and negative sample and utilize the information from hard negative samples, we apply the loss function based on importance sampling. Additionally, we have designed a novel geography encoder that incorporates geographical information and proximity property among positions, allowing us to implicitly capture distance-aware location transitions and spatial clustering phenomenon. We have also utilize proximity-aware negative samplers to promote the informativeness of negative samples. The proposed model PASR is evaluated with three real-world datasets, and the experimental results show that the proposed algorithm outperforms the state-of-the-art sequential location recommendation method significantly. Through the ablation study and sensitivity analysis, we also demonstrate the effectiveness of the loss based on importance sampling, the geography encoder, the grid mapper and the proximity-aware negative sampler at improving recommendation performance.

\begin{adjustwidth}{-\extralength}{0cm}

\reftitle{References}


\bibliography{Ref.bib}

\PublishersNote{}
\end{adjustwidth}
\end{document}